\documentclass[10pt,twocolumn,letterpaper]{article}

\usepackage[pagenumbers]{cvpr} 

\usepackage{multirow}   
\usepackage{floatrow}
\usepackage{amsmath,amsfonts}
\usepackage{algorithmic}
\usepackage{algorithm}
\usepackage{array}
\usepackage{amssymb}
\usepackage{graphicx}
\usepackage{booktabs}
\usepackage{amsthm}
\usepackage{thmtools, thm-restate}
\usepackage{amsmath}
\usepackage{amsfonts}
\usepackage{amssymb}
\usepackage[table,xcdraw]{xcolor}

\usepackage[pagebackref,breaklinks,colorlinks]{hyperref}

\def\paperID{418} 
\def\confName{CVPR}
\def\confYear{2025}

\title{
DimensionX: Create Any 3D and 4D Scenes from a Single Image with \\
Controllable Video Diffusion}

\author{Wenqiang Sun\footnotemark[1]~~$^{1,3}$, Shuo Chen\footnotemark[1]~~$^{2}$, Fangfu Liu\footnotemark[1]~~$^{2}$, Zilong Chen$^{2,3}$, \\  {Yueqi Duan$^2$}, {Jun Zhang\footnotemark[2]~~$^{1}$}, Yikai Wang\footnotemark[2]~~$^{2}$\\
$^{1}$HKUST
~~$^2$Tsinghua University ~~$^3$ShengShu \\
\small \texttt{wsunap@connect.ust.hk}, ~~\texttt{\{chenshuo20,liuff23,chenz122\}@mails.tsinghua.edu.cn},\\
\small \texttt{duanyueqi@tsinghua.edu.cn}, ~~\texttt{eejzhang@ust.hk}, ~~\texttt{yikaiw@outlook.com}, 
}

\begin{document}

\twocolumn[{
\renewcommand\twocolumn[1][]{#1}
\maketitle
     \centering
     \vspace{-0.25cm}
        \includegraphics[width=1.0\linewidth]{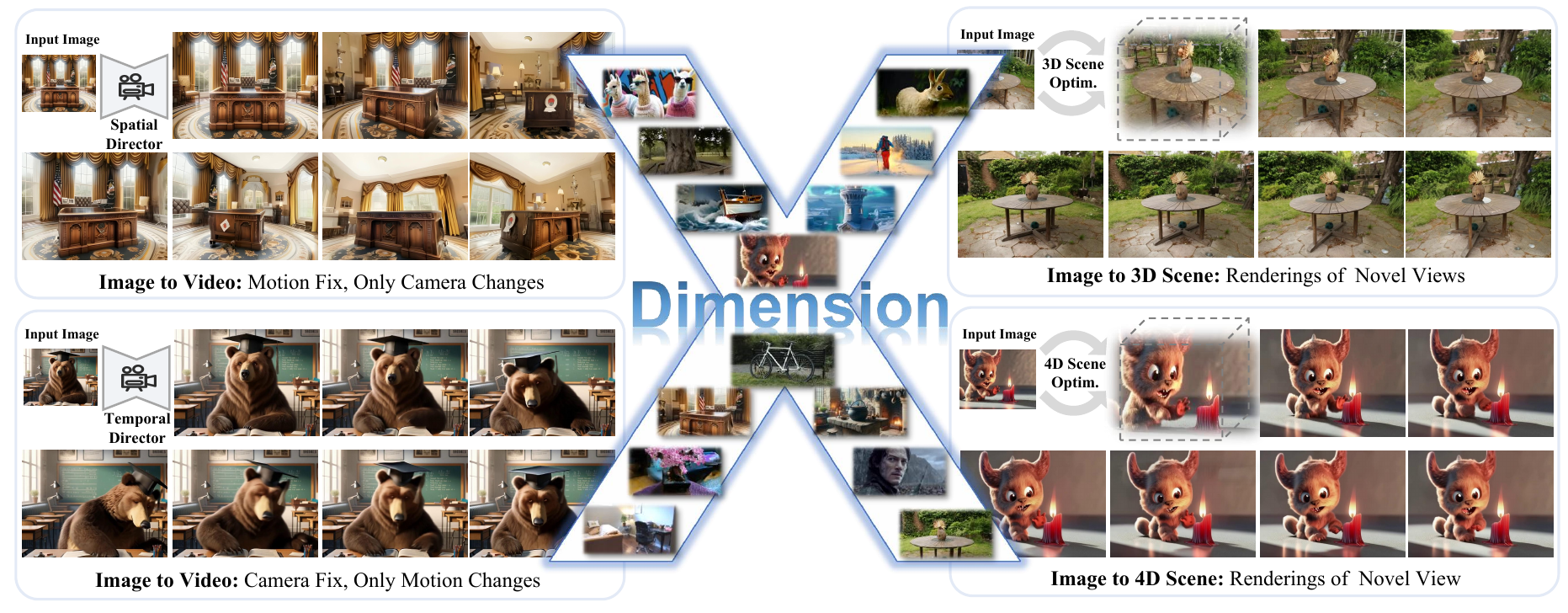}
        \vspace{-0.3cm}
        \captionof{figure}{With just a single image as input, our proposed \textbf{DimensionX} can generate highly realistic videos and 3D/4D environments that are aware of spatial and temporal dimensions.} \vspace{2em}
        \label{fig:intro}
}]
\renewcommand{\thefootnote}{\fnsymbol{footnote}}
\footnotetext[1]{Equal Contribution.  \footnotemark[2]Corresponding author.}

\begin{abstract}
In this paper, we introduce \textbf{DimensionX}, a framework designed to generate photorealistic 3D and 4D scenes from just a single image with video diffusion. Our approach begins with the insight that both the spatial structure of a 3D scene and the temporal evolution of a 4D scene can be effectively represented through sequences of video frames. While recent video diffusion models have shown remarkable success in producing vivid visuals, they face limitations in directly recovering 3D/4D scenes due to limited spatial and temporal controllability during generation. To overcome this, we propose ST-Director, which decouples spatial and temporal factors in video diffusion by learning dimension-aware LoRAs from dimension-variant data. This controllable video diffusion approach enables precise manipulation of spatial structure and temporal dynamics, allowing us to reconstruct both 3D and 4D representations from sequential frames with the combination of spatial and temporal dimensions. Additionally, to bridge the gap between generated videos and real-world scenes, we introduce a trajectory-aware mechanism for 3D generation and an identity-preserving denoising strategy for 4D generation. Extensive experiments on various real-world and synthetic datasets demonstrate that DimensionX achieves superior results in controllable video generation, as well as in 3D and 4D scene generation, compared with previous methods. Project Page: \url{https://chenshuo20.github.io/DimensionX/}
\end{abstract}.    
\section{Introduction}

 In the context of computer graphics and vision, understanding and generating 3D and 4D content are pivotal to create realistic visual experiences \cite{chen2024survey,wu2024recent}. By representing spatial (3D) and temporal (4D) dimensions, videos serve as a powerful medium for capturing dynamic real-world scenes. 
 Despite substantial advancements in 3D and 4D reconstruction technologies \cite{kerbl20233d,Wu_2024_CVPR,wang2024shape,wang2024vidu4d}, there remains a critical shortage of large-scale 3D and 4D video datasets, limiting the potential for high-quality 3D and 4D scene generation from a single image. This scarcity poses a fundamental challenge in constructing photorealistic and interactive environments.

Fortunately, recent advancements in video diffusion models have shown considerable promise in understanding and simulating real-world environments~\cite{blattmann2023stable,yang2024cogvideox}. Driven by advanced video diffusion models, recent works~\cite{voleti2025sv3d,liang2024diffusion4d,xie2024sv4d,yu20244real} have made attempts to leverage the spatial and temporal priors embedded in video diffusion to generate 3D and 4D content from a single image. Despite these rapid developments, existing methods either concentrate on the object-level generation with video diffusion trained on static or dynamic mesh renderings~\cite{voleti2025sv3d,liang2024diffusion4d,xie2024sv4d} or employ time-intensive per-scene optimization for coarse scene-level generation~\cite{yu20244real} (\eg,  Score Distillation Sampling~\cite{poole2022dreamfusion}). This leaves the generation of coherent and realistic 3D/4D scenes an open challenge.

In this paper, we present \textbf{DimensionX}, a novel approach to create high-fidelity 3D and 4D scenes from a single image with controllable video diffusion. While recent video diffusion models are capable of producing realistic results, it remains difficult to reconstruct 3D and 4D scenes directly from these generated videos, primarily due to their poor spatial and temporal controllability during the generation process. Our key insight is to decouple the temporal and spatial factors in video diffusion, allowing for precise control over each individually and in combination. To achieve the dimension-aware control, we establish a comprehensive framework to collect datasets that vary in spatial and temporal dimensions. With these datasets, we present ST-Director, which separates spatial and temporal priors in video diffusion through dimension-aware LoRAs. Additionally, by analyzing the denoising mechanics in video diffusion, we develop a training-free composition method that achieves hybrid-dimension control. With this control, DimensionX generates sequences of spatially and temporally variant frames, enabling the reconstruction of 3D appearances and 4D dynamic motions. To handle complex real-world scenes with our ST-Director, we design a trajectory-aware approach for 3D generation and an identity-preserving denoising mechanism for 4D generation.  
Extensive experiments demonstrate that our DimensionX outperforms previous methods in terms of visual quality and generalization for 3D and 4D scene generation, indicating that video diffusion models offer a promising direction for creating realistic, dynamic environments.

In summary, our main contributions are:
\begin{itemize}
\item We present DimensionX, a novel framework for generating photorealistic 3D and 4D scenes from only a single image using controllable video diffusion.
\item We propose ST-Director, which decouples the spatial and temporal priors in video diffusion models by learning (spatial and temporal) dimension-aware modules with our curated datasets. We further enhance the hybrid-dimension control with a training-free composition approach according to the essence of video diffusion denoising process.
\item To bridge the gap between video diffusion and real-world scenes, we design a trajectory-aware mechanism for 3D generation and an identity-preserving denoising approach for 4D generation, enabling more realistic and controllable scene synthesis.
\item Extensive experiments manifest that our DimensionX delivers superior performance in video, 3D, and 4D generation compared with baseline methods.
\end{itemize}

\begin{figure*}
    \centering
    \includegraphics[width=1\linewidth]{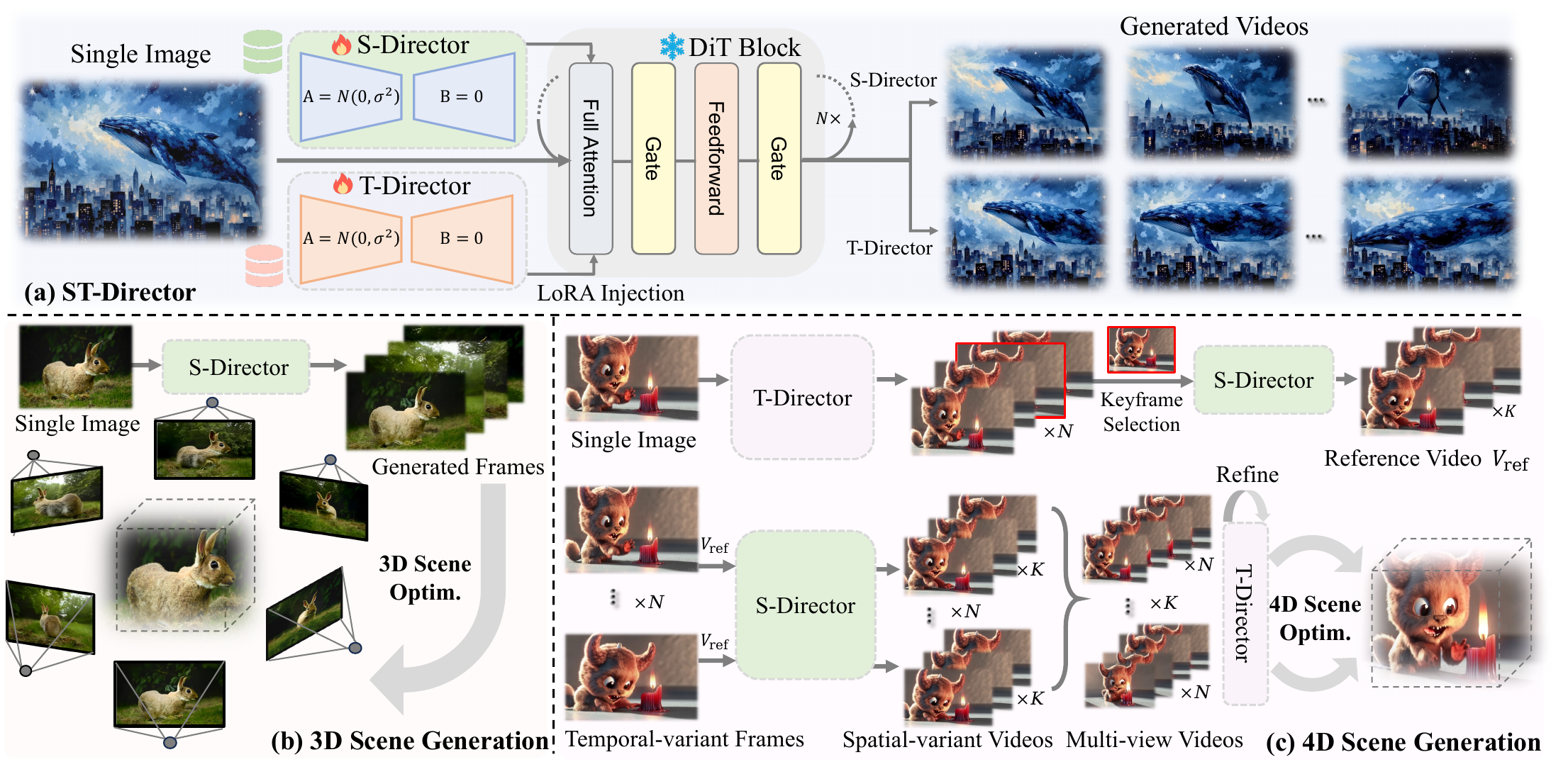}
    \caption{\textbf{Pipeline of DimensionX.} Our framework is mainly divided into three parts. \textbf{(a) Controllable Video Generation with ST-Director.} We introduce ST-Director to decompose the spatial and temporal parameters in video diffusion models by learning dimension-aware LoRA on our collected dimension-variant datasets.  \textbf{(b) 3D Scene Generation with S-Director.} Given one view, a high-quality 3D scene is recovered from the video frames generated by S-Director.  \textbf{(c) 4D Scene Generation with ST-Director.} Given a single image, a temporal-variant video is produced by T-Director, from which a key frame is selected to generate a spatial-variant reference video. Guided by the reference video, per-frame spatial-variant videos are generated by S-Director, which are then combined into multi-view videos. Through the multi-loop refinement of T-Director, consistent multi-view videos are then passed to optimize the 4D scene.} 
    \label{fig:pipeline}
\end{figure*}

\section{Related work}
\label{related_work}

\noindent \textbf{Controllable Video Diffusion.}\quad The integration of additional conditions for conditional image and video generation~\cite{Zhang_2023_ICCV} has seen notable success. Much of the prior research in video generation has focused on injecting various control signals to guide diffusion models. For instance, some approaches control video diffusion using camera pose trajectories, often utilizing ControlNet~\cite{he2024cameractrl,wang2024motionctrl}, Plücker coordinates embeddings~\cite{xu2024camco, bahmani2024vd3d,kuang2024collaborative}, or other coordinate embeddings~\cite{yang2024direct,watson2024controlling}. Other methods leverage reference videos~\cite{ling2024motionclone} or object motion trajectories~\cite{zhang2024tora,wang2024motionctrl} to drive generation. Additionally, several studies have explored other control signals, such as action-based controls~\cite{gao2024vista}, sketches and depth maps~\cite{guo2025sparsectrl}. However, these methods often rely on specific data annotations, which not only limit their scalability but also lead to performance degradation when external control signals are injected into video models. Rather than injecting additional control signals, some methods customize video diffusion models by fine-tuning on a set of reference videos with similar patterns~\cite{guo2023animatediff,zhao2023motiondirector,wei2024dreamvideo}, primarily focusing on Unet-based video diffusion models with spatial and temporal layers. However, this form of customization often lacks generalization. In this work, our method is designed for controlling the DiT-based~\cite{peebles2023scalable} video diffusion model, CogVideoX~\cite{yang2024cogvideox}, which leverages the 3D VAE and full attention to effectively blend spatial and temporal information. Our method eliminates the need for additional signal injection, offering scalability and generalization without sacrificing performance.

\noindent \textbf{3D Generation with Diffusion Priors.}\quad Leveraging 2D diffusion priors for generating 3D content has revolutionized the field of 3D generation. Score Distillation Sampling (SDS) ~\cite{poole2022dreamfusion, wang2024prolificdreamer, lin2023magic3d} distills 2D diffusion priors  to produce high-fidelity 3D meshes from text inputs. To further enhance the 3D consistency, several works have explored the object-level generation by conditioning 2D diffusion models on multi-view camera poses ~\cite{shi2023mvdream, ye2024dreamreward, liu2024sherpa3d, wu2024unique3d, long2024wonder3d}. Similar techniques have been further applied in the scene-level generation ~\cite{sargent2023zeronvs}. More recent approaches leverage video diffusion models to generate novel views from a single image input, achieving impressive results at both the object-level~\cite{voleti2025sv3d, chen2024v3d, gao2024cat3d} and scene-level ~\cite{chen2024v3d, gao2024cat3d, liu2024physics3d}. Additionally, ReconX ~\cite{liu2024reconx} addresses the challenge of sparse-view inputs by employing video interpolation techniques, showcasing the potential of video diffusion models for 3D scene generation. In this work, we unleash the power of video diffusion models in a novel way to generate 3D objects and scenes from a single image input, achieving the competitive performance with fewer training costs and data required.

\noindent \textbf{4D Generation with Diffusion Priors.} Similar to 3D generation, 4D generation has seen significant advancements with the pre-trained diffusion models, including image and video diffusion. Early works \cite{ren2023dreamgaussian4d,bahmani20244d,zhao2023animate124, wang2023animatabledreamer} adopt the SDS technique to per-scene optimize the 4D representation from a text or image input. However, these methods tend to cost hours to generate a 4D asset with obvious inconsistency. To enhance the consistency and generation efficiency, subsequent works \cite{xie2024sv4d,liang2024diffusion4d} filter out high-quality dynamic mesh data from the large-scale Objaverse dataset \cite{deitke2023objaverse,deitke2024objaverse}. Based on these dynamic meshes, a large number of multi-view videos are rendered to train a multi-view video diffusion model. Although these models can generate high-quality 4D multi-view videos, they mainly focus on the object-centric setting rather than the complex scene. For the generation of 4D scenes, the lack of sufficient data poses significant challenges in producing multi-view videos that are utilized to reconstruct the whole scene. 
More recently, 4Real \cite{yu20244real} firstly proposes distilling the pre-trained video diffusion prior with the SDS loss to produce a photorealistic dynamic scene. Unlike the aforementioned works, our approach emphasizes generating temporally and spatially decomposed videos, which are subsequently merged to create multi-view videos for high-quality 4D scene reconstruction.

\section{Methodology}

Given a single image, our goal is to generate high-quality 3D and 4D scenes with controllable video diffusion. To achieve the effective control in respect of spatial and temporal dimensions, we first develop a systemic framework to build the dimension-variant datasets (Sec. \ref{Sec:data_engine}). With the curated datasets, we introduce ST-Director to decouple the spatial and temporal basis through the dimension-aware lora, allowing the precise dimension-aware control. Furthermore, we explore the denoising mechanism during the video generation process and introduce a training-free dimension-aware composition for effective hybrid-dimension control (Sec. \ref{Sec:controllable_video}). To better leverage controllable video diffusion for generating high-quality scenes, we design a trajectory-aware mechanism for 3D generation and an identity-preserving denoising approach for 4D generation (Sec. \ref{Sec:3dgen} and Sec. \ref{Sec:4dgen}).


\subsection{Building Dimension-variant Dataset}\label{Sec:data_engine}

To decouple spatial and temporal parameters in video diffusion, we introduce a framework to collect spatial- and temporal-variant videos from open-source datasets. Notably, we employ a trajectory planning strategy for spatial-variant data and flow guidance for temporal-variant data.


\noindent \textbf{Trajectory planning for spatial-variant data.}
To acquire the spatial-variant dataset, we propose reconstructing photorealistic 3D scenes and rendering videos consistent with our spatial variations. 
To facilitate the selection and planning of rendering paths, we need to compute the coverage range of the cameras throughout the entire scene. Given $N$ cameras in a scene, we first compute the center $C$ and principal axes $A$ along the direction x, y, and z using the Principal Component Analysis (PCA) technique:
\begin{equation}
      C = \frac{\sum_{i=1}^{N} \boldsymbol{p}_i}{N},\quad  A = \text{SVD}(\mathcal{P} - C),
\end{equation}
where $p_i$ denotes the position of camera $i$, $\mathcal{P}=\left\{\boldsymbol{p}_i, 1 \leq i \leq N\right\} \in \mathbb{R}^{N \times 3}$ represents the position set of $N$ cameras, and $\text{SVD}$ is the Singular Value Decomposition operation. Next, we need to calculate the lengths $L$ of each axis from the projection distance $D$:
\begin{align}
    D &= (\mathcal{P} - C) \cdot A \\
      L &= \text{max}(D) - \text{min}(D).
\end{align}

Built on the above calculation, we have already figured out the distribution of the camera throughout the entire scene.
 To cope with various scenes, we establish the following rules to filter out the qualifying data: \textbf{(1) Camera Distribution}: We calculate the center of the scene and judge how cameras capture around the scene. \textbf{(2) Bounding Box Aspect Ratio}: The aspect ratio of the bounding box should meet the requirement for various S-Directors. For instance, the aspect ratio of $x$ and $y$ axis should not vary too greatly, which helps in selecting appropriate 360-degree surrounding videos.
\textbf{(3) Camera-to-Bounding Box Distance}: We calculate the distance from each camera position to the closest plane of the bounding box and prioritize data with smaller total distances to ensure better camera placement.

With the filtered dataset, it is necessary to compute the occupancy field within the scene to help us plan the feasible region for the rendering cameras. After reconstructing the entire scene’s 3DGS from multi-view images, we render multi-view images and their corresponding depth maps, and then use TSDF \cite{curless1996volumetric} to extract the scene’s mesh from the RGB-D data. More details can be found in appendix.



\noindent \textbf{Flow guidance for temporal-variant data.}  To achieve the temporal control, we aim to filter the temporal-variant data to fine-tune the video diffusion model. Specifically, we leverage the optical flow \cite{teed2020raft} to filter out the temporal-variant videos. For temporal-variant videos, optical flow maps frequently exhibit extensive white regions, which could serve as an effective selection criterion.



\begin{figure*}
    \centering
    \includegraphics[width=1.0\linewidth]{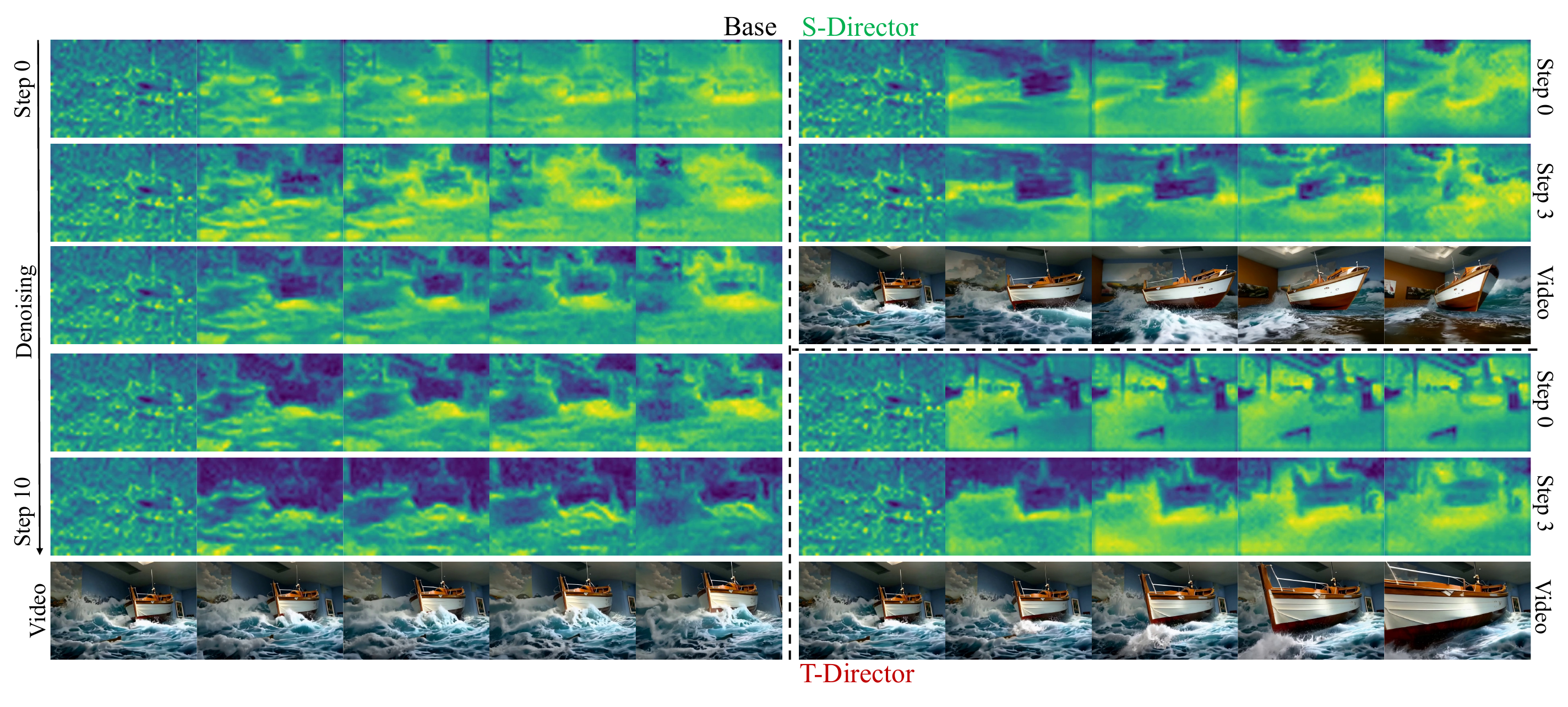}
    \caption{\textbf{Visualization of Attention Map.} The left row shows the attention maps during the denoising process of the original video diffusion model. The right row, from top to bottom, illustrates the attention map variation of S-Director and T-Director, respectively. Starting from step 0, the early denoising steps (before step 10 of total denoising step 50) have determined the outline and layouts of output videos. Specifically, the spatial component is recovered earlier than the temporal information during the denoising process.}
    
    \label{fig:attn_map}
\end{figure*}

\subsection{ST-Director for Controllable Video Generation}\label{Sec:controllable_video}

Inspired by the concept of orthogonal decomposition in linear algebra, we propose a method to decouple spatial and temporal dimensions in video generation for more precise control. We conceptualize each video frame \( I_t(u, v) \) as a projection from a 4D space \(\mathbb{R}^3 \times \mathbb{R}^1\), where \( u \) and \( v \) are the image coordinates in the frame, and the 4D space consists of three spatial dimensions \( x, y, z \) and a temporal dimension \( t \). In this framework, the 4D space consists of a static background \( B \subset \mathbb{R}^3 \) and dynamic objects \( O^i(t) \subset \mathbb{R}^3 \), where \( i \) indexes the moving objects at time \( t \). The entire scene at time \( t \) is then represented as:
\begin{align}
\mathcal{S}(t) = B \cup \left( \bigcup_i O^i(t) \right).
\end{align}

Each video frame at time \( t \) is, therefore, a 2D projection of this 3D scene structure onto the image plane, governed by the camera’s parameters at that moment. To formalize this, we define the projection function \( \mathcal{P}_{C(t)} \), which projects the 3D scene \( \mathcal{S}(t) \) onto a 2D image:
\begin{align}
I_t(u, v) = \mathcal{P}_{C(t)}(\mathcal{S}(t)),
\end{align}
where \( C(t) \) represents the camera parameters at time \( t \). 

However, this reduction from 4D to 2D is typically treated as an inseparable blend of spatial and temporal dimensions in most video generation methods, complicating independent control over each dimension, which makes its difficult to recover the 4D space. To address this, we introduce two orthogonal basis directors: the \textbf{S-Director} (Spatial Director) and the \textbf{T-Director} (Temporal Director), which allow us to separate spatial and temporal variations within the video generation process. With these orthogonal basis directors, we can more flexibly control video generation, generating frames along a single axis or combining both directors to achieve projections that represent any desired perspective within the 4D space. 

\subsubsection{Dimension-aware Decomposition}\label{Sec:decomposition_lora}

To decompose spatial and temporal variations, we define the spatial and temporal equivalence relations that capture the behavior of points in the 4D space under different conditions. More details can be found in our appendix. From these two equivalence relations, we derive two types of quotient spaces, \( \mathbb{R}^4 / \sim_S \) and \( \mathbb{R}^4 / \sim_T \). The S-Quotient Space, \( \mathbb{R}^4 / \sim_S \), captures the spatial trajectory of camera viewpoints in a video by collapsing the temporal dimension, as each frame represents different spatial perspectives at a single point in time. Conversely, the T-Quotient Space, \( \mathbb{R}^4 / \sim_T \), describes temporal object motion trajectories from a fixed camera position, collapsing the spatial dimension in terms of viewpoint, with the primary variation being temporal as objects move within the scene. Together, these two quotient spaces, representing spatial and temporal decompositions of the 4D space, enable us to interpret video as a structured decomposition of 4D space into distinct spatial and temporal components.

Building on these quotient spaces, we associate videos from our spatial-variant dataset with the S-Quotient Space, while videos from our temporal-variant dataset correspond to the T-Quotient Space.  In order to train the S-Director and T-Director to generate videos specifically within these spatial and temporal structures, we employ LoRA \cite{hu2021lora}, a fine-tuning method that is both parameter-efficient and computationally light, training each director separately on one of the two datasets to guide the video diffusion model. Specifically, the S-Director is trained on the spatial-variant dataset, learning patterns in which time is held constant (\( \mathcal{S}(t) = \mathcal{S}_0 \)), thereby generating videos within the S-Quotient Space, as illustrated in the top right of Fig. \ref{fig:attn_map}, where \( I_t(u, v) = \mathcal{P}_{C(t)}(\mathcal{S}_0) \). Similarly, the T-Director is trained on the temporal-variant dataset, learning patterns where the camera remains stationary (\( C(t) = C_0 \)), resulting in videos within the T-Quotient Space, as shown in the bottom right of Fig. \ref{fig:attn_map}, with \( I_t(u, v) = \mathcal{P}_{C_0}(\mathcal{S}(t)) \). 

\subsubsection{Tuning-free Dimension-aware Composition}\label{Sec:composition_lora}

With this orthogonal basis of directors, we achieve flexible control over video generation, where each director independently captures frame sequences along its designated axis, producing either \( I_t(u, v) = \mathcal{P}_{C_t}(\mathcal{S}(0)) \) or \( I_t(u, v) = \mathcal{P}_{C_0}(\mathcal{S}(t)) \), respectively. However, most videos naturally involve a blend of spatial and temporal elements, making it essential to combine both directors to capture richer, multifaceted perspectives within the 4D space, represented as \( I_t(u, v) = \mathcal{P}_{C_t}(\mathcal{S}(t)) \). To achieve this enhanced level of control, we aim to merge the S-Director and T-Director, allowing for dynamic adjustment that aligns video generation with specific spatial, temporal, or combined spatiotemporal intents. In pursuit of this goal, we examine the underlying mechanics both of the base model’s and each director's denoising process by visualizing the attention maps generated by the base model alongside both directors (as shown in Fig. \ref{fig:attn_map}). These visualizations reveal two key observations:

\noindent \textbf{Observation 1}: \textit{The initial steps of the denoising process are critical for defining the generated video.}

From the attention maps, we observe that during the initial steps of the denoising process, both the base model and the two directors establish foundational sketches that closely align with the final generated results. These preliminary outlines capture essential spatial and temporal structures early on, effectively setting the direction for the remaining denoising steps. Additionally, we notice a distinct difference in how these changes unfold: in the base model, both temporal and spatial alterations occur simultaneously, creating a unified evolution across both dimensions. In contrast, when utilizing the S-Director and T-Director, only one dimension changes at a time, either temporal or spatial, depending on the director. 





\noindent \textbf{Observation 2}: \textit{Spatial information is constructed earlier than temporal information.}

Similar to findings from Motionclone~\cite{ling2024motionclone}, we observe that the synthesis of object motion remains underdeveloped in the early stages of the denoising process. Specifically, with S-Director, the attention maps reveal that the structural outlines of the final video appear much earlier than with temporal control.  As evidence, Fig. \ref{fig:attn_map} shows that at the same step 0 and 3 of the denoising loop, the object remains stationary with the T-Director, while the S-Director already guides the camera to move through the scene.

\begin{figure*}[h]
    \centering
    \includegraphics[width=1\linewidth]{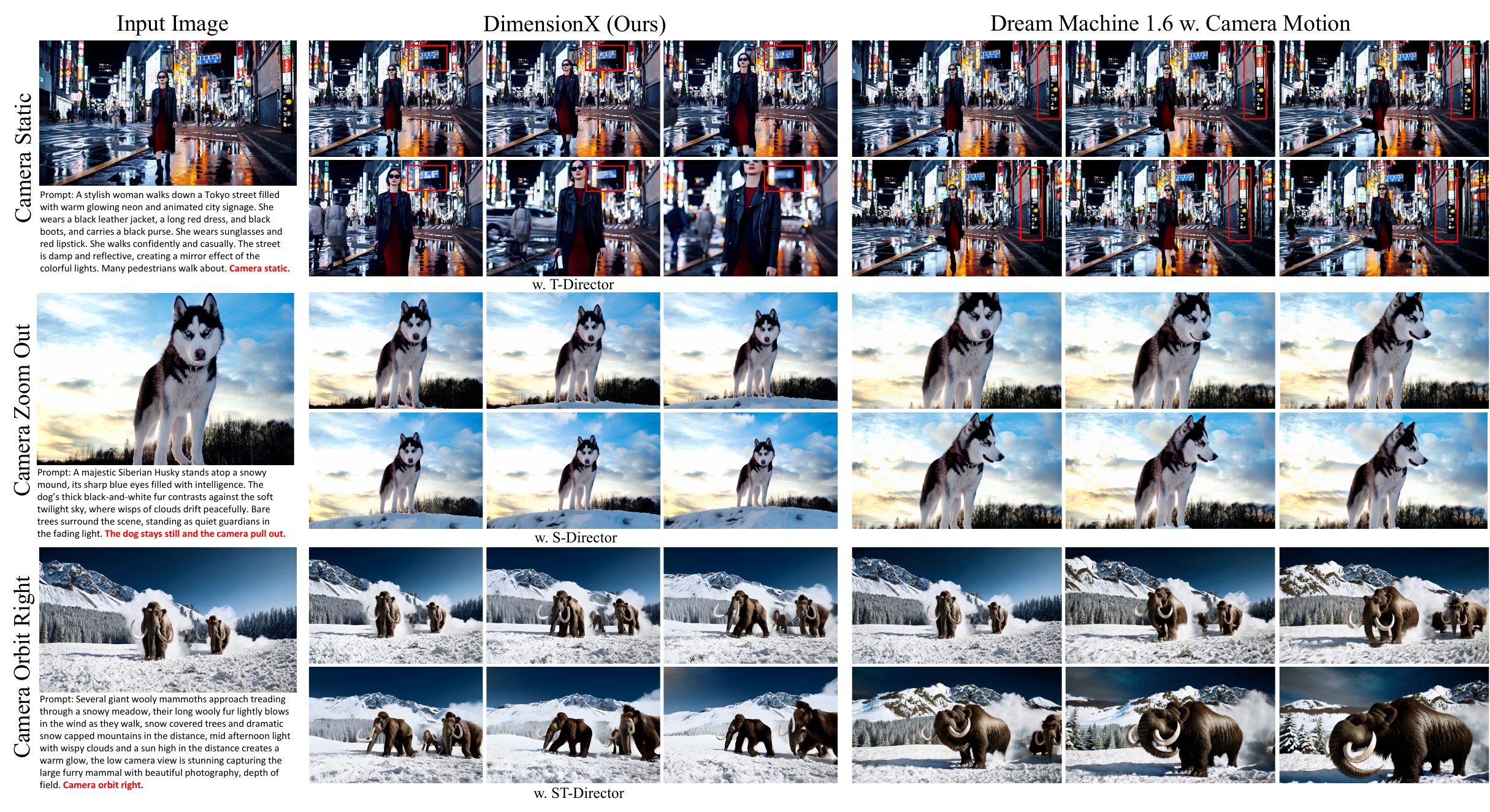}
    \caption{\textbf{Qualitative comparison in dimension-aware video generation.} Given the same image and text prompt, the first row is the temporal-variant video generation (camera static), the second row is the  spatial-variant video generation (camera zoom out), and the third row is the spatial- and temporal-variant video generation (camera orbit right).}
    \label{fig:compare_video}
\end{figure*}

Based on the two observations above, we propose a training-free method, \textbf{Switch-Once}, a novel approach to compose diverse LoRAs. This approach combines the S-Director and T-Director to generate videos that seamlessly blend spatial and temporal information, achieving a balanced synthesis represented by \( I_t(u, v) = \mathcal{P}_{C_t}(\mathcal{S}(t)) \). Following Observation 2, we initiate the denoising process with the S-Director to establish comprehensive camera motion across the scene. Then, as indicated by Observation 1, we switch to the T-Director after the initial steps of the denoising process (our experiments show that transitioning at the 4th or 5th step yields optimal results), thereby enhancing the quality of object motion in the final video. The resulting video, as shown in Fig. \ref{fig:compare_video}, demonstrates the effectiveness of this balanced approach. 

\subsection{3D Scene Generation with S-Director} \label{Sec:3dgen}

Built upon the S-Director, our video diffusion model is able to generate long-range consistent frames from a single image, allowing for the reconstruction of photorealistic scenes.  To better generalize to real-world scenarios, where spatial variations are diverse and camera trajectories are highly flexible, we introduce a \textbf{trajectory-aware mechanism} to handle different camera movements. 
Specifically, to cover a wide range of camera trajectory patterns \(C(t)\), we train multiple types of S-Directors, each tailored to a specific camera motion. In the 3D world, camera movements are defined by 6 degrees of freedom (DoF), with each DoF allowing movement in both positive and negative directions for translation and rotation, resulting in 12 distinct motion patterns. Additionally, we also train orbital motion category S-Director, where the camera follows a smooth, circular path around the subject, capturing a unique perspective beyond the standard DoF-based movements. With diverse and controllable S-Directors, we adopt the trajectory-aware mechanism in both single-view and sparse-view settings, enabling generalizable generation of real-world 3D scenes.

\noindent \textbf{Single-view Scene Generation.} Given a single image $I$, our goal is to reconstruct the 3D scene with generated video frames $\left\{{I}^i\right\}_{i=1}^{N}$, where $N$ represents the frame length. Although current video diffusion models have shown potential for long video generation, the total duration still falls far short of the frame count required for real-world scene reconstruction. Specifically, the powerful open-source video diffusion model (\eg CogVideoX \cite{yang2024cogvideox}) currently generates a maximum of only 49 frames, whereas reconstructing a large scene (\eg 360 degree scene) typically requires hundreds of multi-view images.
To address this, we extend the video diffusion model to generate 145 frames (three times the original frame count). 



\noindent \textbf{Sparse-view Scene Generation.} In applications where the accuracy and detail in 3D scene generation are paramount, using sparse view inputs can significantly enhance the fidelity of the generated content. In this setting, we propose incorporating a video interpolation model and an adaptive S-Director to achieve a smooth and consistent transition between the sparse views.
First, we develop a video diffusion model to generate the high-quality interpolated video, which takes two images as the start and end frames.   
Specifically, given two-view images $\left\{I^1, I^2\right\}$, we concatenate the noisy latent $z_2 = \mathcal{E}(I^2)$ to the ending of sequential latents, mirroring the common practice of concatenating the first frame’s latent $z_1 = \mathcal{E}(I^1)$ with its noisy counterpart. The objective function for the video diffusion process is formulated as

\begin{equation}
\mathcal{L}_\text{diffusion} = \mathbb{E}_{z_t \sim p, \epsilon \sim \mathcal{N}(0, I), t} \left[ \|\epsilon - \epsilon_{\theta}({z}_t, t, z_1, z_2, c)\|_2^2 \right],
\end{equation}
where $z_t$ is the noisy latent sequence from training videos, and \(\epsilon_{\theta}\) represents the model's prediction of the noise at timestep \(t\), conditioned on the first and last frame latent: \(z_1\) and \(z_2\). With the interpolated video diffusion model, we then train various S-Directors to provide refined camera motion guidance, ensuring smooth and consistent transition between the sparse-view images. In particular, we tailor two key strategies to fully leverage the guidance prior carried in S-Directors: early-stopping training and adaptive trajectory-planning. Our findings reveal that training the S-Director to an early phase sufficiently achieves robust trajectory guidance, where the camera movement can be flexibly modulated with changes in input viewpoints. Furthermore, to deal with various viewpoint interpolations of spare views, we propose adaptively selecting the appropriate S-Director according to the coordinate relation between the input images. More details can be found in Appendix.

Powered by our proposed long-video diffusion model and diverse S-Directors, extensive scenes can be directly reconstructed from the generated videos. In particular, given sparse-view (\ie as few as one) images and a chosen camera motion type, which can be a basic trajectory or a combination of these motion primitives, our video diffusion can generate a consistent long video along the specified path. 
To alleviate the inconsistency behind generated videos, we adopt an confidence-aware gaussian splatting procedure to reconstruct the 3D scene. Initialized with the point cloud and estimated camera poses from DUSt3R \cite{wang2024dust3r}, 3D gaussian splatting is optimized with additional LPIPS loss~\cite{zhang2018unreasonable} and confidence maps acquired from DUSt3R. We adopt the 3DGS loss as follow:
\begin{equation}
\begin{aligned}
\mathcal{L}_{\text {conf }}= \mathcal{C} \left(\lambda_{\text{1}} \mathcal{L}_1 + \lambda_{\text{ssim}} \mathcal{L}_{\text{ssim}} + \lambda_{\text{lpips}} \mathcal{L}_{\text{lpips}} \right),
\end{aligned}
\end{equation}
where $\mathcal{C}$ is the confidence maps, and $\lambda_{\text{1}}, \lambda_{\text{ssim}}, \lambda_{\text{lpips}}$ represent the coefficients. More details can be found in appendix.

\begin{figure*}[t]
    \centering
    \includegraphics[width=1\linewidth]{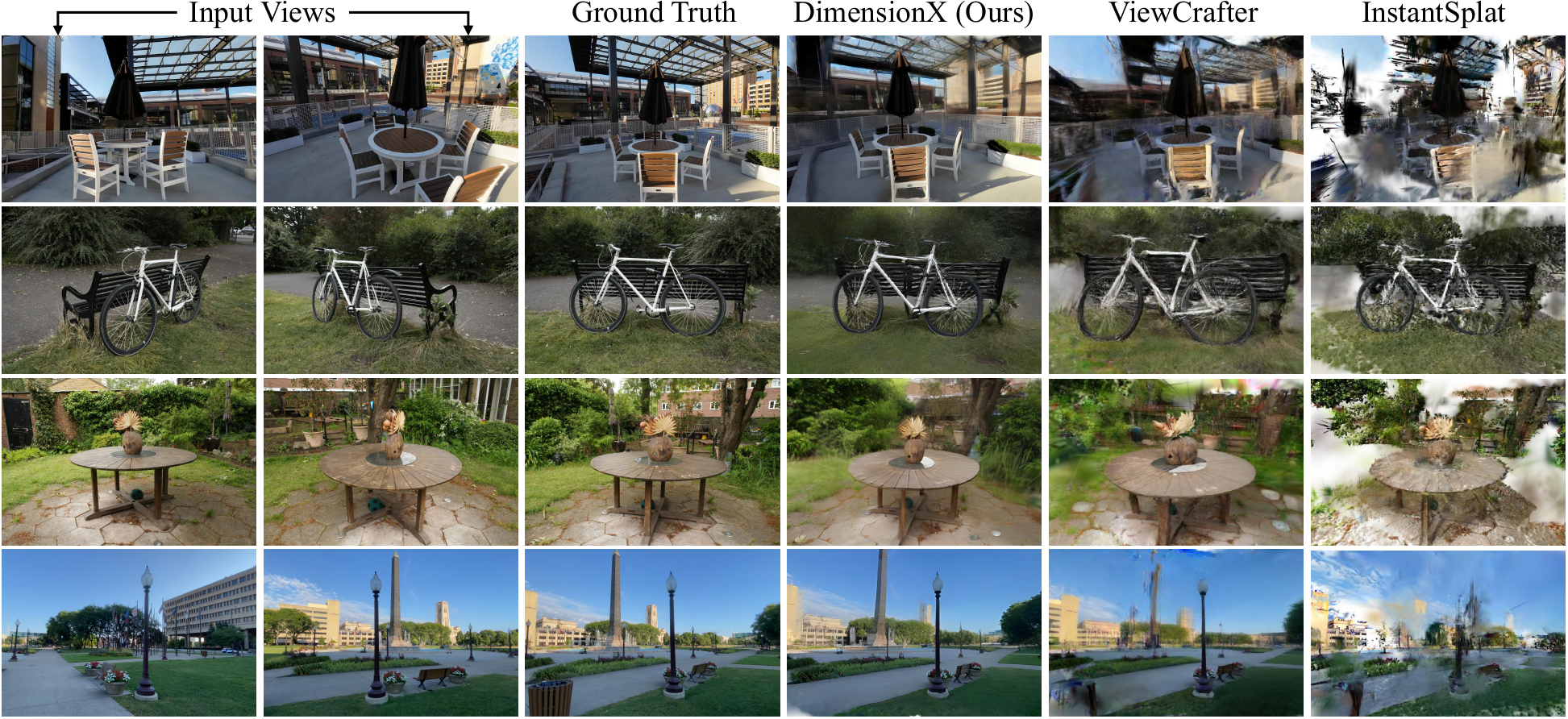}
    \caption{\textbf{Qualitative Comparison in sparse-view 3D generation.} Given two large-angle views, our approach obviously outperforms other baselines.}
    \label{fig:sparse_compare}
\end{figure*}

\subsection{4D Scene Generation with ST-Director} \label{Sec:4dgen}
Equipped with spatial and temporal controlled video diffusion, we can recover a high-quality 4D dynamic scene from a single image. A direct way is to stitch together the spatial-variant videos generated for each frame in temporal-variant videos into multi-view videos, which are then used to reconstruct the 4D scene. However, this method has a key challenges: \textbf{3D consistency}. 
Maintaining consistency in the background and object appearance across spatial-variant videos is challenging, causing severe jitter and discontinuity in the 4D scene.
To address the above issue, we propose an \textbf{identity-preserving denosing} strategy, including the reference video latent sharing and appearance refinement process, to enhance the consistency of all spatial-variant videos.

Given an input image ${I}$, our goal is to generate a photorealistic 4D scene with dynamic objects and high-quality backgrounds. First, we employ T-Director to generate a temporal-variant video frames $\left\{{I}^i\right\}_{i=1}^{N}$ for the input image, from which a reference frame $I_{\textit{ref}}$ is selected to produce corresponding spatial-variant video frames $v_{\textit{ref}}= \left\{{I}^i\right\}_{i=1}^{K}$, where $K$ represents the number of cameras. Subsequently, $v_{\textit{ref}}$ is used to guide the generation of spatial-variant videos across all temporal-variant video frames $\left\{{I}^i\right\}_{i=1}^{N}$, which are then combined into multi-view videos $\left\{ \left\{{I}^i_j\right\}_{i=1}^{N} \right\}_{j=1}^{K}$. $\left\{{I}^i_j\right\}_{i=1}^{N}$ represents the temporal-variant video captured from the camera $j$. Despite leveraging the reference video to guide the generation process, minor shape inconsistencies still exist, causing temporal jitter or inter-view misalignment. To mitigate these issues, we introduce an extra appearance refinement process to further enforce consistency across the multi-view video.
With consistent multi-view videos, we choose the deformable 3D Gaussian Splatting \cite{Wu_2024_CVPR} to model the dynamic scene. 

\noindent \textbf{Reference Video Latent Sharing.}\quad Through our empirical study, we propose choosing the reference frame based on the dynamic object’s mask size and the magnitude of optical flow values, allowing us to acquire a frame that best encompasses the dynamic object. With the chosen reference frame $I_{\textit{ref}}$, S-Director is applied to produce the corresponding spatial-variant video $v_{\textit{ref}}$. Applying the forward diffusion process on $v_{\textit{ref}}$, we derive the noisy latent code ${z}_{\textit{ref}}$ as following:
\begin{equation}
{z}_{\textit{ref}}=\sqrt{{\alpha}_t} {z}_0+\sqrt{1-{\alpha}_t} \epsilon, \quad \epsilon \sim \mathcal{N}(\mathbf{0}, \mathbf{1}),
\label{forward}
\end{equation}
where $z_0 = \mathcal{E}(v_{\textit{ref}})$, representing the compressed latent by the encoder $\mathcal{E}$, and $\sqrt{{\alpha}_t}$ determines the strength of using the reference video.
Starting from the same initialization latent ${z}_{\textit{ref}}$, all frames are subsequently denoised to produce spatial-variant videos with strong coherence. Moreover, to further enhance consistency among these frames, we propose blending the denoised  $z_t$  of each frame with the corresponding reference video latent $z_{\textit{ref},t}$ at the early denoising steps:
\begin{equation}
    z_t = \lambda z_t + (1-\lambda) z_{\textit{ref},t} ,
\end{equation}
where $\lambda$ is an adjustable hyperparameter.

\noindent \textbf{Appearance Refinement.}\quad To further enhance the stability and consistency between the combined multi-view videos, the appearance refinement process is applied to the dynamic video from each viewpoint. Inspired by the image-to-image translation SDEdit \cite{meng2021sdedit}, we apply random noise to each multiview video $v_j=\left\{{I}^i_j\right\}_{i=1}^{N}$, and perform multi-step denoising, acquiring smooth and high-quality videos with the video diffusion prior: 
\begin{equation}
v^{\text {refine }}_j=f_\theta\left(v_j+\epsilon\left(t_{0}\right) ; t_{0}, c\right),
\end{equation}
where $t_0$ represents the forward diffusion timestep, and $v^{\text {refine }}_j$ is the refined video with the denoise function $f_\theta$ of T-Director. In addition, we repeat the refine process during the middle timestep to enhance the smoothness.

Having acquired consistent multiview videos $\left\{ \left\{{I}^i_j\right\}_{i=1}^{N} \right\}_{j=1}^{K}$, we use the deformable 3dgs \cite{Wu_2024_CVPR} to reconstruct the 4d scene. Following the previous reconstruction methods, we apply the $L_1$, total-variational, and ssim loss to optimize the scene:
\begin{equation}
\begin{aligned}
\mathcal{L}= \mathcal{L}_1 + \mathcal{L}_{\text{tv}} + \mathcal{L}_{\text{ssim}}.
\end{aligned}
\end{equation}



\begin{table*}[t]
\small
\caption{\textbf{Quantitative comparison of single-view and sparse-view scenarios.} Our approach outperforms other baselines in all metrics both in terms of single-view and sparse-view settings.}
\label{tb:3d result}
\centering

\resizebox{\linewidth}{!}{
    \begin{tabular}{c*{13}{c}}
        \toprule
        & \multirow{2}{*}{\textbf{Methods}} & \multicolumn{3}{c}{Tank and Temples} & \multicolumn{3}{c}{MipNeRF360} & \multicolumn{3}{c}{LLFF} & \multicolumn{3}{c}{DL3DV} \\
        \cmidrule(lr){3-5} \cmidrule(lr){6-8} \cmidrule(lr){9-11} \cmidrule(lr){12-14}
        & & {PSNR $\uparrow$} & {SSIM} $\uparrow$ & {LPIPS} $\downarrow$ & {PSNR $\uparrow$} & {SSIM} $\uparrow$ & {LPIPS} $\downarrow$ & {PSNR $\uparrow$} & {SSIM} $\uparrow$ & {LPIPS} $\downarrow$ & {PSNR $\uparrow$} & {SSIM} $\uparrow$ & {LPIPS} $\downarrow$ \\
        \midrule
        \multirow{3}{*}{Single-View}& ZeroNVS \cite{sargent2023zeronvs} & \cellcolor{yellow!20}12.31& \cellcolor{yellow!20}0.301 & \cellcolor{yellow!20}0.567 & \cellcolor{orange!20}15.84 & \cellcolor{yellow!20}0.327 & \cellcolor{yellow!20}0.536 & \cellcolor{yellow!20}15.62 & \cellcolor{yellow!20}0.497 & \cellcolor{yellow!20}0.354 & \cellcolor{yellow!20}12.39 & \cellcolor{yellow!20}0.251 & \cellcolor{yellow!20}0.559 \\
        & ViewCrafter \cite{yu2024viewcrafter} & \cellcolor{orange!20}15.18& \cellcolor{orange!20}0.499 & \cellcolor{orange!20}0.319& \cellcolor{yellow!20}15.65 & \cellcolor{orange!20}0.404& \cellcolor{orange!20}0.378 & \cellcolor{orange!20}17.56& \cellcolor{orange!20}0.620& \cellcolor{orange!20}0.337& \cellcolor{orange!20}14.78 & \cellcolor{orange!20}0.422 & \cellcolor{orange!20}0.417 \\
        & \textbf{Ours} & \cellcolor{red!20}17.11 & \cellcolor{red!20}0.613 & \cellcolor{red!20}0.199 & \cellcolor{red!20}18.91 & \cellcolor{red!20}0.527 & \cellcolor{red!20}0.333 &\cellcolor{red!20}20.38 & \cellcolor{red!20}0.744& \cellcolor{red!20}0.200 &\cellcolor{red!20}18.28 & \cellcolor{red!20}0.642& \cellcolor{red!20}0.215 \\
        \midrule
        \multirow{4}{*}{Sparse-View} & DNGaussian \cite{li2024dngaussian} & 12.13 & 0.292 & 0.511 & 15.21 & 0.127 & 0.632 & 17.51 & 0.586 & 0.409 & 14.99 & 0.286 & 0.432 \\
        & InstantSplat \cite{fan2024instantsplat} & \cellcolor{yellow!20}18.70&\cellcolor{yellow!20}0.634 &\cellcolor{yellow!20}0.258 &\cellcolor{yellow!20}16.80 & \cellcolor{yellow!20}0.574& \cellcolor{yellow!20}0.296& \cellcolor{orange!20}22.33& \cellcolor{yellow!20}0.818& \cellcolor{orange!20}0.149& \cellcolor{yellow!20}18.30 & \cellcolor{orange!20}0.691 & \cellcolor{yellow!20}0.222\\
        & ViewCrafter \cite{yu2024viewcrafter} & \cellcolor{orange!20}18.76 & \cellcolor{orange!20}0.637 & \cellcolor{orange!20}0.216 & \cellcolor{orange!20}18.49 & \cellcolor{orange!20}0.691 & \cellcolor{orange!20}0.212 & \cellcolor{yellow!20}21.60 & \cellcolor{orange!20}0.823 & \cellcolor{yellow!20}0.155 & \cellcolor{orange!20}19.19 & \cellcolor{yellow!20}0.686 & \cellcolor{orange!20}0.196 \\
        & \textbf{Ours} & \cellcolor{red!20}20.42 & \cellcolor{red!20}0.668 & \cellcolor{red!20}0.185 & \cellcolor{red!20}20.21 & \cellcolor{red!20}0.713 & \cellcolor{red!20}0.184 & \cellcolor{red!20}25.11 & \cellcolor{red!20}0.913 & \cellcolor{red!20}0.067 & \cellcolor{red!20}21.69 & \cellcolor{red!20}0.780 & \cellcolor{red!20}0.124 \\
        \bottomrule
    \end{tabular}
}

\end{table*}

\section{Experiment}

In this section, we conduct extensive experiments on real-world and synthetic datasets to assess the controllability of our DimensionX, as well as its 3D and 4D scene generation capabilities using ST-Director. We begin with a comprehensive illustration about our experimental details (Sec. \ref{setup}). Then, in Sec. \ref{Sec:controllable_video}, we provide both quantitative and qualitative evaluation in controllable video generation. Following that, we report the quantitative and qualitative results of our approach in comparison to other baselines under various scenarios, including single-view and sparse-view 3D generation (Sec. \ref{3d}). Subsequently, we present the 4D scene generation results of our approach (Sec. \ref{4d}). Finally, we conduct various ablation studies to evaluate the effectiveness of our design (Sec. \ref{ablation}).

\subsection{Experimental Setup} \label{setup} 

\noindent \textbf{Implementation Details.} 
We choose the open-source I2V model CogVideoX \cite{yang2024cogvideox}, which adopts the diffusion transformer architecture, as our video diffusion model. For the ST-Director training, we set the LoRA rank to 256, and finetune the LoRA layers with 3000 steps at the learning rate 1e-3. To enlarge the video frames, we first modify the RoPE~\cite{su2024roformer} positional embedding to extend the video length to 145 frames and decrease the resolution to $480\times320$. For the training of video interpolation models, we first full fine-tune the base model for 2,000 steps at the learning rate 5e-5, then we train the S-Director using the same LoRA configuration but for only 1,000 steps. 
During the inference stage, we adopt the DDIM sampler \cite{song2020denoising} with classifier-free guidance \cite{ho2022classifier}. In 3DGS optimization stage, we select the first and end frames of generated videos to produce the initialization point cloud and 49 frames from videos to optimize the scene. We follow the original 3DGS pipeline \cite{kerbl20233d}, and set the optimization step to 7000 steps. The hyperparameters $\lambda_{\text{1}}$, $\lambda_{\text{ssim}}$, and $\lambda_{\text{lpips}}$ are 0.8, 0.2, and 0.3, respectively. During the deformable 3DGS phrase, we use the generated 32-view videos as the training data, and apply the original deformable 3DGS to reconstruct the 4D scene.

\noindent \textbf{Datasets.} In our whole framework, our video diffusion model is mainly trained on three datasets: DL3DV-10K \cite{ling2024dl3dv}, OpenVid \cite{nan2024openvid}, and RealEstate-10K \cite{zhou2018stereo}. OpenVid-1M \cite{nan2024openvid} is a curated high-quality open-sourced video dataset, including 1 million video clips with diverse motion dynamics and camera controls. DL3DV-10K \cite{ling2024dl3dv} is a widely-collected 3D scene dataset with high-resolution multi-view images, including diverse indoor and outdoor scenes. RealEstate-10K is a dataset from youtube, mianly including the captures of indoor scenes. Applying our designed data collection framework, we build the dimension-variant dataset from DL3DV-10K and OpenVid. We select 100 high-quality temporal-variant videos from OpenVid to train T-Director. For each S-Director type, 100 videos are rendered according to the specific camera trajectory to train the corresponding LoRA. To enlarge the video frame, we filter high-quality videos exceeding 145 frames from the RealEstate-10K \cite{zhou2018stereo} and OpenVid \cite{nan2024openvid} to full fine-tune the video diffusion model. With the same dataset, we fine-tune a video interpolation model with the first and end frame guidance. To further verify the 3D generation ability of our DimensionX, we compare our approach with other baselines on Tank-and-Temples \cite{10.1145/3072959.3073599}, MipNeRF360 \cite{barron2022mip}, NeRF-LLFF \cite{mildenhall2019local}, and DL3DV-10K~\cite{ling2024dl3dv}. 

\begin{figure*}[h]
    \centering
    \includegraphics[width=1\linewidth]{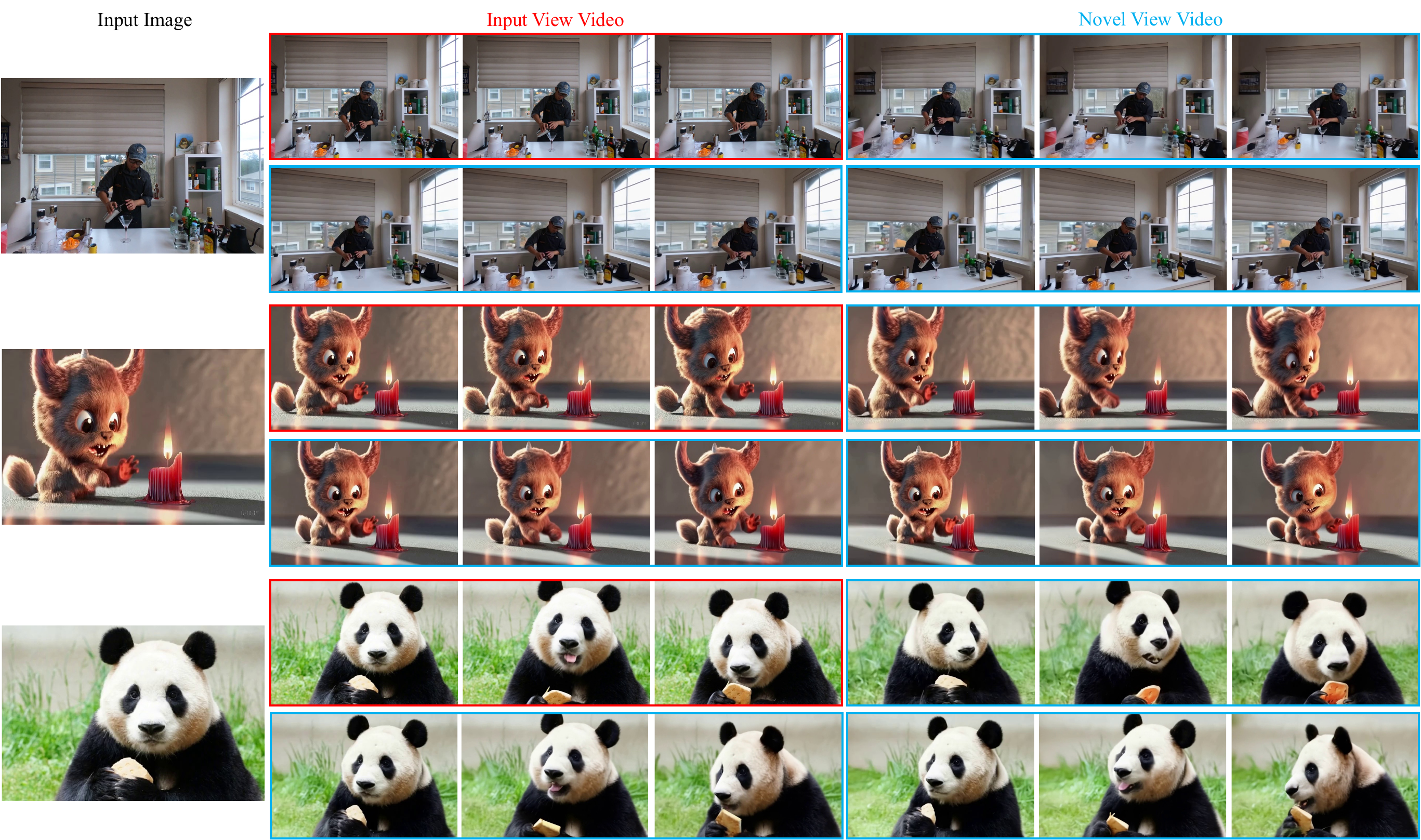}
    \caption{\textbf{Qualitative results of 4D scene generation.} Given a real-world or synthetic single image, our DimensionX produces coherent and intricate 4D scenes with rich features. }
    \label{fig:compare_4d}
\end{figure*}

\subsection{Controllable Video Generation.}

\begin{table}[t]
\footnotesize
    \centering
    \begin{tabular}{cccc}
        \toprule
         & Consistency$\uparrow$ & Dynamic $\uparrow$ & Aesthetic $\uparrow$ \\
         \midrule
         CogVideoX~\cite{yang2024cogvideox} & 93.56 & 11.76 & 57.81 \\
        Dream Machine 1.6 & 93.69 & 38.24 & 68.96\\
        \textbf{Ours} & \textbf{97.69} & \textbf{47.06} & \textbf{70.82}\\
         \bottomrule
    \end{tabular}
    \caption{\textbf{Qualitative comparison for controllable video generation.} Our DimensionX outperforms baseline models in all metrics, including the Consistency, Dynamic, and Aesthetic scores.}
    \label{tab:video}
\end{table}

\noindent \textbf{Baselines and Evaluation Metrics.} We compare our DimensionX with the original CogVideoX~\cite{yang2024cogvideox}(open-source) and Dream Machine 1.6 (closed-source product). We collect hundreds of images as our evaluation dataset. Following the previous benchmark VBench~\cite{huang2024vbench}, we evaluate the Subject Consistency, Dynamic Degree, and Aesthetic Score of generated videos as our metrics.

\noindent \textbf{Quantitative and Qualitative Comparison.} 
The qualitative result in Table \ref{tab:video} demonstrates the impressive performance of our approach, including the better visual quality and 3D consistency. 
As shown in Fig. \ref{fig:compare_video}, we can observe that our DimensionX achieves effective decomposition of spatial and temporal parameters of video diffusion model, while Dream Machine cannot decouple the dimension-aware control, even though we utilize the camera motion and prompt constraint. Moreover, for the hybrid-dimension control, including spatial and temporal motion, in comparison to Dream Machine, our DimensionX generates more impressive and dynamic videos. Both quantitative and qualitative results indicate that our approach can create controllable videos while maintaining the dynamic motions and subject consistency.

\subsection{3D Scene Generation.} \label{3d}

\noindent \textbf{Baselines and Evaluation Metrics.} In the single-view setting, we compare our approach with two generative methods: ZeroNVS \cite{sargent2023zeronvs} and ViewCrafter \cite{yu2024viewcrafter}. For the sparse-view scenario, we select two sparse-view reconstruction methods and one sparse-view generation baseline, including: DNGaussian \cite{li2024dngaussian}, InstantSplat \cite{fan2024instantsplat}, and ViewCrafter \cite{yu2024viewcrafter}. We adopt PSNR, SSIM, and LPIPS as the metric for our quantitative results. Specifically, In both single-view and sparse-view settings, we begin by reconstructing the 3D scene from the given images, followed by calculating the metrics using renderings from novel views.

\noindent \textbf{Quantitative and Qualitative Comparison.}
The quantitative comparison results are presented in Table \ref{tb:3d result}. We can observe that DimensionX outperforms the baselines in all metrics, demonstrating the impressive performance of our approach. As presented in Fig. \ref{fig:sparse_compare}, in both single-view (More details can be found in our appendix.) and sparse-view settings, our approach can reconstruct high-quality 3D scenes, while other baselines fail to handle the challenging cases.

\subsection{4D Scene Generation} \label{4d}


We evaluate our DimensionX on both real-world and synthetic datasets. Specifically, we adopt the Neu3D \cite{li2022neural}, which contains high-resolution multi-view video of different scenes, to verify the performance of our approach in real-world video to 4D generation. As shown in Fig. \ref{fig:compare_4d},
given a single image, our DimensionX generates highly consistent dynamic videos from large-angle novel views.

\subsection{Ablation Study} \label{ablation}

\begin{figure*}
    \centering
    \includegraphics[width=1\linewidth]{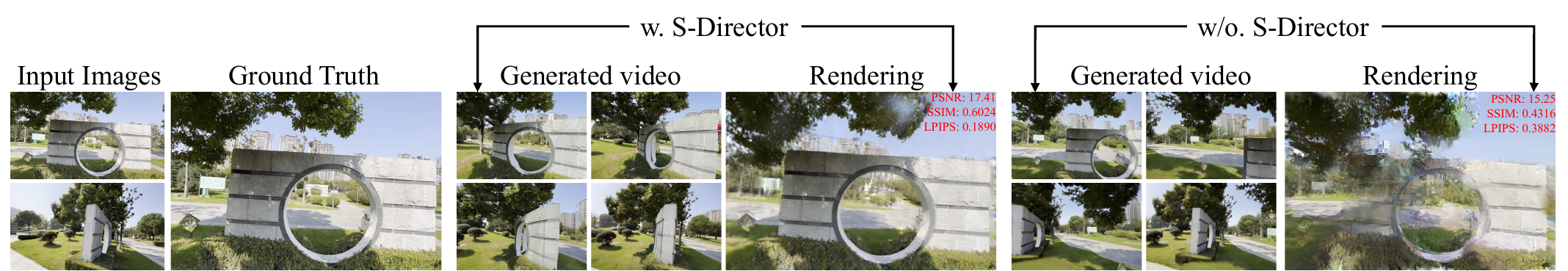}
    \caption{\textbf{Ablation study on the sparse-view 3D generation}: The absence of S-Director results in lower reconstruction quality.}
    \label{fig:ablation}
\end{figure*}

\noindent \textbf{Trajectory-aware mechanism for 3D generation.} In sparse view 3D generation, we leverage S-Director to guide video interpolation model. As illustrated in Fig. \ref{fig:ablation}, when handling the large-angle sparse view, the absence of S-Director often results in the "Janus problem", where multiple heads are generated, significantly degrading reconstruction quality.

\begin{figure}[h]
    \centering
    \includegraphics[width=1\linewidth]{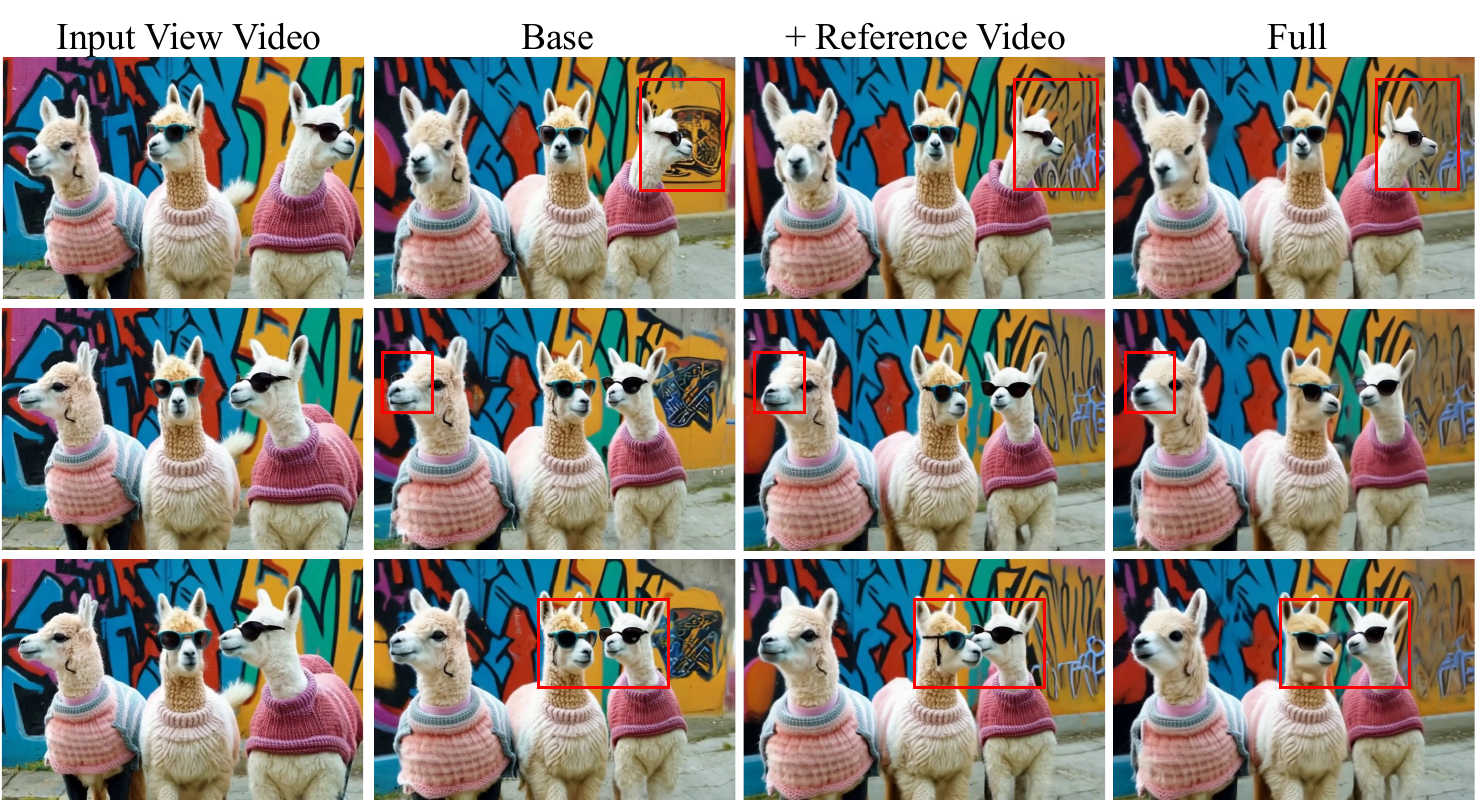}
    \caption{\textbf{Ablation study on 4D generation} We ablate the design of reference video latent sharing and appearance refinement.}
    \label{fig:ablation_4d}
\end{figure}

\noindent \textbf{Identity-preserving denoising strategy for 4D generation.} In 4D scene generation, we conduct experiments on real-world images to analyze our identity-preserving denoising for 4D scene generation. As shown in Fig. \ref{fig:ablation_4d}, we ablate the design of reference video latent sharing and appear refinement in terms of the consistency among different frames of a novel view. Specifically, we can observe that directly combing per-frame videos causes severe inconsistency, including the background and subject shape. Through the reference video latent sharing, the global background and appearance exhibit a high consistency across different frames. Building on reference video latent sharing, appearance refinement enhances the coherence of appearance details.

\section{Conclusion}
In this paper, we introduce DimensionX, a novel framework to create photorealistic 3D and 4D scenes from only a single image with controllable video diffusion. Our key insight is to introduce ST-Director to decouple the spatial and temporal priors in video diffusion models by learning the dimension-aware LoRA on dimension-variant datasets. Furthermore, we investigate the denoising process of video diffusion and introduce a tuning-free dimension-aware composition to achieve the hybrid-dimension control. Powered by the controllable video diffusion, we can recover accurate 3D structures and 4D dynamics from the sequential generated video frames. To further enhance the generalization of our DimensionX in real-world scenes, we tailor a trajectory-aware strategy for 3D scene generation and an identity-aware mechanism for 4D scene generation. Extensive experiments on various real-world and synthetic datasets demonstrate that our approach achieve the state-of-the-art performance in controllable video generation, as well as 3D and 4D scene generation.

\noindent \textbf{Limitations and future work.} Despite the remarkable achievements, our DimensionX is limited by the diffusion backbone. Although current video diffusion models are capable of synthesizing vivid results, they still struggle with understanding and generating subtle details, which restricts the quality of the synthetic 3D and 4D scenes. Additionally, the prolonged inference procedure of video diffusion models hampers the efficiency of our generation process. In the future, it is worthy to explore how diffusion models can be integrated for more efficient end-to-end 3D and 4D generation. We believe that our research provides a promising direction to create a dynamic and interactive environment with video diffusion models.

{
    \small
    \bibliographystyle{ieeenat_fullname}
    \bibliography{main}
}


\end{document}